\documentclass[preprint,3p,times]{elsarticle}

\usepackage{graphicx}
\usepackage{epstopdf, epsfig}
\usepackage{amsmath}
\usepackage{subcaption}
\usepackage{bm}
\usepackage{upgreek} 

\usepackage{ifthen}
\usepackage{xcolor}

\usepackage[linesnumbered,ruled,vlined]{algorithm2e}

\usepackage{graphicx}
\usepackage{amssymb}

\usepackage{lineno}

\biboptions{comma, sort&compress}

\usepackage{bm}
\usepackage{hyperref}
\usepackage{cleveref}
\newcommand{\doc}{paper}
 
\newcommand{\apriori}{\textit{a priori}}

\newcommand{\nwidth}{n--width}
\newcommand{\ndimensional}{n--dimensional}

\newcommand{\stateparnull}{\ensuremath{ w }} 
\newcommand{\stateparc}{\ensuremath{ \stateparnull }} 
\newcommand{\statepar}{\ensuremath{ \bm{\stateparc} }}

\newcommand{\statedis}[1]{\ensuremath{ \bm{\stateparnull} }} %
\newcommand{\stateparinit}{\ensuremath{\stateparnull_{0}}}
\newcommand{\stateparinterp}{\ensuremath{\bm{\stateparnull}^{*}}}

\newcommand{\residualnull}{\ensuremath{ R }}
\newcommand{\residual}[1]{\ensuremath{ \residualnull\left(#1\right) }}
\newcommand{\residualxnull}{\ensuremath{ \residualnull_{x} }}
\newcommand{\residualwnull}{\ensuremath{ \residualnull_{\stateparnull} }}
\newcommand{\minimize}{\ensuremath{ arg\text{min} }}

\newcommand{\realsetnull}{\ensuremath{ \mathbb{R} } } 
\newcommand{\realsetO}[1]{\ensuremath{ \realsetnull^{#1} } } 
\newcommand{\realset}[2]{\ensuremath{ \realsetO{#1 \times #2} }} 

\newcommand{\Norm}[1]{\ensuremath{  \left\| #1 \right\| }}
\newcommand{\Normvec}[1]{\ensuremath{  \left| #1 \right| }}

\newcommand{\NormT}[1]{\ensuremath{  \Norm{#1}_2 }}
\newcommand{\NormvecT}[1]{\ensuremath{  \Normvec{#1}_2 }}

\newcommand{\transpose}[1]{\ensuremath{ {#1}^{\top} }}

\newcommand{\manifold}{\ensuremath{ \mathcal{M} }}
\newcommand{\manifoldS}{\ensuremath{ \mathcal{S} }}
\newcommand{\reducem}[1]{\ensuremath{ \reducef{\bm{#1}}_{n} }}
\newcommand{\reducef}[1]{\ensuremath{ \widetilde{  #1 }     }}

\newcommand{\numgrid}{\ensuremath{N_{x}}}

\newcommand{\numtime}{\ensuremath{N_{t}}}

\newcommand\Rey{\mbox{\textit{Re}}}  
\newcommand{\viscosity}{\ensuremath{ \nu }} 
\newcommand{\convspeed}{\ensuremath{ c }} 

\newcommand{\loss}{\ensuremath{\mathcal{\bm{L}}}}
\newcommand{\lossresidual}{\ensuremath{ \loss_{r} }}
\newcommand{\lossresidualx}{\ensuremath{ \loss_{{r}_{x}} }}
\newcommand{\lossresidualw}{\ensuremath{ \loss_{{r}_{w}} }}
\newcommand{\lossbc}{\ensuremath{  \loss_{bc} }}
\newcommand{\lossic}{\ensuremath{  \loss_{ic} }}

\newcommand{\activation}{\ensuremath{\phi}}
\newcommand{\activationi}[2]{\ensuremath{ {\activation}_{#1}\left({#2}\right) }}

\newcommand{\weight}[1]{\ensuremath{ \bm{W}_{#1} }}
\newcommand{\NNparam}{\ensuremath{ \bm{\theta} }}

\newcommand{\numlayer}{\ensuremath{m}}
\newcommand{\numwidth}{\ensuremath{w}}
\newcommand{\numstate}{\ensuremath{\Dim}}

\newcommand{\biasi}[1]{\ensuremath{ \bm{b}_{#1}}}

\newcommand{\transformlayer}[1]{\ensuremath{\gamma\left({#1}\right)}}

\newcommand{\inputvec}{\ensuremath{ \bm{u} }}

\newcommand{\nntheta}[1]{\ensuremath{ \mathcal{N} \left(#1\right) }}
\newcommand{\nnthetax}[1]{\ensuremath{ \mathcal{N}_x \left(#1\right) }}
\newcommand{\nnthetau}[1]{\ensuremath{ \mathcal{N}_{\stateparnull} \left(#1\right) }}

\newcommand{\tmax}{\ensuremath{ T }}

\newcommand{\penaltyr}{\ensuremath{ \lambda_r }}
\newcommand{\penaltyic}{\ensuremath{ \lambda_{ic} }}

\newcommand{\ic}{\ensuremath{ \textsl{g} }}

\newcommand{\penaltybc}{\ensuremath{ \lambda_{bc} }}
\newcommand{\bc}{\ensuremath{ \mathcal{B} }}

\newcommand{\Dim}{\ensuremath{ {d} }}

\newcommand{\Nic}{\ensuremath{ N_{ic} }}

\newcommand{\Nbc}{\ensuremath{ N_{bc} }}
\newcommand{\Nr}{\ensuremath{ N_{r} }}

\newcommand{\fracsum}[1]{\ensuremath{ \frac{1}{#1} \sum_{i=1}^{#1} }}

\usepackage{amsopn}

\usepackage[acronym]{glossaries}
\newacronym{rvm}{RVM}{relevance vector machine}
\newacronym{svd}{SVD}{singular value decomposition}
\newacronym{ml}{ML}{machine learning}
\newacronym{dnn}{DNN}{deep neural network}
\newacronym{hit}{HIT}{Homogeneous Isotropic Turbulence}
\newacronym{ale}{ALE}{arbitrary Lagrangian--Eulerian}

\newacronym{pinn}{PINN}{physics--informed neural network}
\newacronym{xpinn}{XPINN}{extended \gls{pinn}}
\newacronym{lpinn}{LPINN}{Lagrangian physics--informed neural network}
\newacronym{pde}{PDE}{partial differential equation}
\newacronym{nn}{NN}{neural network}
\newacronym{lstm}{LSTM}{Long short--term memory}
\newacronym{ann}{ANN}{artificial neural network}
\newacronym{rom}{ROM}{reduced order model}
\newacronym{lspg}{LSPG}{least--square Petrov--Galerkin}

\newacronym{pid}{PID}{principal interval decomposition}
\newacronym{tvd}{TVD}{total variation diminishing}
\newacronym{alm}{ALM}{Augmented Lagrangian method}

\newacronym{ntk}{NTK}{Neural Tangent Kernel}
\newacronym{1d}{1D}{one--dimensional}
\newacronym{2d}{2D}{two--dimensional}

\newcommand{\myscale}{1.00}
\newcommand{\myscalearch}{1.3}
\def\myscalethree{1.00}
\newcommand{\mycolor}{black}
\usepackage{makecell}
\def\mydepository{\url{https://github.com/rmojgani/LPINNs}}

\journal{}

\begin{document}
	
\begin{frontmatter}
		

\title{Lagrangian PINNs: \\
	A causality--conforming solution to failure modes\\
	of physics-informed neural networks
}


%

\author[add1]{Rambod Mojgani}
\ead{rm99@rice.edu}
\author[add2]{Maciej Balajewicz}
\author[add1,add3]{Pedram Hassanzadeh}

\address[add1]{Department of Mechanical Engineering, Rice University, Houston, TX}
\address[add2]{Realtor.com, Santa Clara, CA}
\address[add3]{Department of Earth, Environmental and Planetary Sciences, Rice University, Houston, TX}

\begin{abstract}
Physics--informed neural networks (PINNs) leverage neural--networks to find the solutions of partial differential equation (PDE)--constrained optimization problems with initial conditions and boundary conditions as soft constraints.
These soft constraints are often considered to be the sources of the complexity in the training phase of PINNs.
Here, we demonstrate that the challenge of training 
(i) persists even when the boundary conditions are strictly enforced, and 
(ii) is closely related to the Kolmogorov n--width associated with problems demonstrating transport, convection, traveling waves, or moving fronts.
Given this realization, we describe the mechanism underlying the training schemes such as those used in eXtended PINNs (XPINN), curriculum regularization, and sequence--to--sequence learning. 
For an important category of PDEs, i.e., governed by non--linear convection--diffusion equation,
we propose reformulating PINNs on a Lagrangian frame of reference, i.e., LPINNs, as a PDE--informed solution. 
A parallel architecture with two branches is proposed. 
One branch solves for the state variables on the characteristics, and the second branch solves for the low--dimensional characteristics curves.
The proposed architecture conforms to the causality innate to the convection, and leverages the direction of travel of the information in the domain.
Finally,  we demonstrate that the loss landscapes of LPINNs are less sensitive to the so--called ``complexity'' of the problems, compared to those in the traditional PINNs in the Eulerian framework.
\end{abstract}


\end{frontmatter}


\section{Introduction}
\label{sec:introduction}
The evolution of many physical phenomena and engineering systems can be derived from first principles leading to governing equations in the form of \glspl{pde}.
Although analytical solutions of many of non--linear \glspl{pde} are seldom known, development of numerical methods has made approximation of the solution possible. 
One of the paradigms of solving for a state of the system is through optimization. 
A solution satisfies the governing \gls{pde},
\begin{equation}\label{eq:residual}
\residual{ \stateparnull\left(x,t\right)} := \frac{\partial \stateparnull\left(x,t\right) }{\partial t} - \mathcal{N}\left( \stateparnull\left(x,t\right)\right)=0,
\end{equation}
where  $\stateparnull\left(x,t\right)$ is the state parameter on a spatial domain of $x\in\Omega$ and $t\in\left[0,T\right]$ with appropriate boundary and initial conditions.
Therefore, finding a solution of a \gls{pde}, $\stateparnull^*$, is equivalent to finding a minimizer of the residual equation, i.e.,

\begin{equation}\label{Eqn:ResMin}
\begin{aligned}
\stateparnull^* =
& \underset{\stateparnull}{~\minimize}~
\residual{ \stateparnull\left(x,t\right)},
\end{aligned}
\end{equation}
subject to boundary and initial conditions as constraints.
Iterative methods are traditionally used to 
find minimizers of high--dimensional non--linear residual equations.

On the other hand, in the absence of the governing equations, where the phenomena/task cannot be described using first principles, \gls{ml} methods such as \glspl{ann} are recognized to be compelling.
Therefore, the application of \glspl{ann} to solve the systems with known governing equations seem to be superfluous.
Nevertheless, in recent years, these paradigms have merged in \glspl{pinn}~\cite{Raissi_JCP_2019}, such that \glspl{ann} are trained to find the minimizer of \gls{pde}--constrained optimization.

Despite the maturity of the developed numerical methods of solving \glspl{pde}, the flexibility in implementation, readily available adjoints via automatic differentiation, and the low inference cost of \glspl{pinn} have made it an appealing tool~\cite{Karniadakis_nature_2021}, especially in inverse problems and inverse design~\cite{Lu_SISC_2021}, ill--posed/conditioned problems~\cite{Arzani_POF_2021}, and control~\cite{Mowlavi_arxiv_2021}.

However, the training phase of \glspl{pinn}, equivalent to solving the~\glspl{pde}, faces some challenges~\cite{Fuks_JMLMC_2020, Mao_JCP_2020, Wang_JCP_2022, Wang_SISC_2021, Krishnapriyan_NIPS_2021, Bihlo_JCP_2021, Wang_arxiv_2022, Basir_SCITECH_2022,Colby_CCP_2020}.
The innovations and attempts to improve the accuracy of the \glspl{pinn} can be classified into two categories.
In the first category, the \gls{ann} architecture and loss are targeted to improve the training behavior.
In~\cite{Wang_JCP_2022}, it is shown that the eigenvalues of the \glspl{ntk} of different loss components explains the training behavior.
Accordingly, penalty weights in the loss function are adaptability determined at each iteration of the training. 
Similarly in \cite{Wang_SISC_2021}, the unbalanced gradients of the components of the loss is associated with training failure, and annealing the learning rate is proposed. 
It is also demonstrated that the architecture of the network can meaningfully change the stiffness of the gradients in the learning phase, and therefore it is suggested that a specialized architecture can be beneficial to specific problems~\cite{Wang_SISC_2021}.
Subsequently, reformulating the constraints using \gls{alm} \cite{Hestenes_JOTA_1969}
demonstrates a flatter/smoother loss landscape, and therefore leads to a more favorable training behavior \cite{Basir_SCITECH_2022}, compared to the originally proposed penalty terms \cite{Raissi_JCP_2019}.
Notably, the loss landscape of \glspl{pinn} are less smooth compared to purely data--driven \glspl{ann} \cite{Krishnapriyan_NIPS_2021, Basir_SCITECH_2022}.
The raggedness of loss landscape explains why  \glspl{pinn} are more prone to converge to an unfavorable local minima.
Such challenge depends on both the governing equations, and system parameters \cite{Fuks_JMLMC_2020, Krishnapriyan_NIPS_2021, Basir_SCITECH_2022}.

In the second category, prior knowledge/property of the system is enforced.
For instance, in hyperbolic systems, e.g., inviscid Burgers' equation, \gls{tvd}, and entropy inequalities can be imposed in addition to artificial viscosity \cite{Patel_JCP_2022}. 
In \cite{Fuks_JMLMC_2020}, the flux term of non--linear convection diffusion had to be modified to help with the accuracy of the solution.
However, such solutions are \gls{pde} specific, and are not applicable to all the challenging test cases.
Approaches such as curriculum learning (training on a simple problem and transferring the learned weights to the harder problems)~\cite{Krishnapriyan_NIPS_2021},
adaptive sampling (in both space and time) \cite{Colby_CCP_2020}, or a very similar sequence--to--sequence learning (decomposing the temporal domain)~\cite{Krishnapriyan_NIPS_2021} can be applied to a wider range of problems without any prior assumptions.
Note that sequence--to--sequence learning~\cite{Krishnapriyan_NIPS_2021} or adaptive sampling \cite{Colby_CCP_2020} are different than of the parallel--in--time decomposition of the temporal domain, where different networks, each corresponding to one of the temporal subdomains, are trained~\cite{Bihlo_JCP_2021}.
In many of the aforementioned studies, the same network is shown to be capable of expressing a more accurate solution, given additional considerations in the loss and training. Such experiments demonstrate that the used architectures are expressive enough, and therefore, the challenge lies in the training regime. 
It is argued that all such challenges can be more naturally overcome by respecting the underlying spatio--temporal \textit{causality}~\cite{Wang_arxiv_2022}.
One natural approach to impose such causality is by prioritizing the earlier time steps in the training phase \cite{Wang_arxiv_2022}. 

Moreover, the type of the differential operators, i.e., parabolic, hyperbolic, or elliptic, is believed to describe the difficulty in discovery of the minima. 
In the case of elliptic and parabolic \glspl{pde}, the generalization error, i.e., the difference between a global minimizer of the loss and the solution to the \glspl{pde}, converges to zero under certain conditions, given enough number of data points~\cite{Shin_CCP_2020}.
However, similar results are lacking for hyperbolic equations.
More importantly, the optimization error, i.e., the difference between the global and local minima given some data points, is poorly understood.

\Cref{tab:studies} summarizes challenging test cases and the proposed remedies. 
While chaotic systems, e.g., turbulent flow, and Lorenz 63, and higher order derivatives in governing equations, 
are known to be challenging since the earliest formulation of \glspl{pinn}~\cite{Wang_arxiv_2022}, they merely cannot explain the difficulty in training of non--chaotic systems, e.g., advection/convection, reaction, reaction-diffusion, Poisson's, or wave equations. 

\begin{table}[!tb]
	{\footnotesize
		\caption{
			Identified challenging problems in the training of \glspl{pinn}.
		}\label{tab:studies}
		\begin{center}
			\begin{tabular}{|c|c|c|c|c|} \hline
				\bf Study 
				& \bf \glspl{pde}
				& \bf Domain							
				& \bf Test case 			
				& \bf Proposed remedy 
				\\ 
				\hline
				\cite{Fuks_JMLMC_2020} 			
				& Convection--diffusion
				& 1D
				& Buckley--Leverett 		
				& Introducing viscosity 
				\\
				\hline
				\cite{Colby_CCP_2020}
				& \makecell{
					Reaction--diffusion\\
					Adsorption/\\ desorption --\\ surface diffusion
				}
				& \makecell{1D \\ 2D \\ 3D}
				& \makecell{Allen-Cahn \\ Cahn-Hilliard}
				&
				\makecell{
					Time adaptive \\ (sampling/marching),\\
					and mini--batching,\\ 
					and regularizing the \\ loss components
				}
				\\
				\hline
				\cite{Mao_JCP_2020}				
				& Euler
				& 1D 
				& Sod's shock tube			
				& \makecell{Characteristic form,\\ Oversampling the shock} \\
				\hline
				\cite{Fraces_arxiv_2021} 
				& Convection--diffusion
				& 1D
				& Buckley--Leverett 		
				& Tuning the flux term
				\\
				\hline
				\cite{Wang_SISC_2021} 			
				&  \makecell{Helmholtz\\
					Klein–Gordon\\
					Navier– Stokes
				}
				& \makecell{2D\\1D\\ 2D}
				& \makecell{N/A\\N/A\\ Lid--driven cavity\\ ($\Rey=100$)}
				& \makecell{Annealing \\ the learning rate}
				\\
				\hline
				\cite{Bihlo_JCP_2021} 			
				& \makecell{Advection\\
					Shallow-water}
				& \makecell{Spherical}
				& {\color{\mycolor}\makecell{
						Traveling feature
				}}
				& \makecell{Parallel in \\time decomposition}
				\\
				\hline
				\cite{Krishnapriyan_NIPS_2021} 	
				&\makecell{
					Convection\\
					Reaction\\
					Reaction-diffusion
				}
				&\makecell{1D}
				&\makecell{ Challenging \\in specific regimes}
				&\makecell{Curriculum \\ regularization, or \\
					Sequence--to--sequence\\ learning
				}
				\\
				\hline
				\cite{Abreu_ICCS_2021}
				& \makecell{
					Inviscid Burgers'\\
					Convection--diffusion
				}
				& \makecell{1D}
				& \makecell{Shock \\ Rarefaction and \\ shock}
				&
				\makecell{
					Enriching the data--set,\\ 
					and artificial viscosity
				}
				\\
				\hline
				\cite{Wang_JCP_2022} 			
				& Wave 
				& 1D
				& {\color{\mycolor}\makecell{
						Traveling wave
				}}
				& Adaptive penalty
				\\
				\hline	
				\cite{Wang_arxiv_2022} 			
				& \makecell{
					Lorenz 63\\
					Kuramoto--Sivashinsky\\
					Navier--Stokes
				}
				& \makecell{\realsetO{3}\\1D\\2D}
				& \makecell{N/A\\Regular and chaotic\\Decaying turb-\\ulence ($\Rey=100$)}
				& \makecell{
					Causal training of loss,\\ and/or
					transformer \\ architecture
				} 
				\\
				\hline
				\cite{Basir_SCITECH_2022}
				& \makecell{
					Poisson's
				}
				& \makecell{1D \\ 2D}
				& \makecell{High wave--number\\ forcing}
				&
				\makecell{
					Enforcing the \\
					constraints by \gls{alm}
				}
				\\
				\hline
				\cite{Patel_JCP_2022}		
				& \makecell{
					Euler \\
					Convection--diffusion
				} 
				& \makecell{
					1D
				}  
				& \makecell{
					Sod's shock tube \\
					Buckley--Leverett 
				}
				&  
				\makecell{
					\gls{tvd}, and\\
					entropy inequalities}
				\\
				\hline
				{\color{\mycolor}
					Ours}
				& {\color{\mycolor}\makecell{
						Convection--diffusion \\
						Viscous Burgers'
				}}
				& {\color{\mycolor}\makecell{
						1D
				} }
				& {\color{\mycolor}\makecell{
						Traveling feature,\\
						Traveling shock
				}}
				& {\color{\mycolor}\makecell{
						Lagrangian PINN
				}}
				\\
				\hline
			\end{tabular}
		\end{center}
	}
\end{table}

Although the previous studies have described some of the difficulties and dynamics of the training phase, there is no universal theory on convergence rate or \apriori\ measure of success of the training, and the present error estimators need to be tightened for practical non--linear systems~\cite{Hillebrecht_arxiv_2020}.
In this \doc, we provide some evidence that connects the dimensionality of solution, in the Kolmogorov \nwidth\ sense, to the difficulties in the training phase. 
Moreover, we explain how the previous remedies connects to the presented description of complexity. Subsequently, we add an unrecognized complex test case, to the existing list of \cref{tab:studies}, i.e., the Burgers' equation in the presence of a shock sweeping the domain (traveling shock). 
Specifically, we demonstrate that in the cases where the shock sweeps  long distances, the network fails to train, a similar behavior as seen for convection equation~\cite{Krishnapriyan_NIPS_2021}.
{\color{\mycolor}
We emphasize that \gls{pinn} is successfully applied to viscous Burgers' equation where a stationary shock forms~\cite{Raissi_JCP_2019}.
}
 
Finally, we focus on an important category of the recognized challenging cases, i.e., convection--diffusion problems.
We propose \Glspl{lpinn}  to conform to the ``causality'' in the system.
\Glspl{lpinn}' architecture is informed by the direction of travel of information in the domain, i.e., along the characteristic curves.
In this architecture the solution is to be learned with an inherently reduced dimensionality, a simpler task for any of \gls{ml} architectures.

The paper is organized as follows.
In \cref{sec:problem}, some of the challenging cases of training of \glspl{pinn} are discussed.
In \cref{sec:convdiff}, we focus on an important canonical set of problems governed by non--linear convection--diffusion equation, and review the traditional architecture of \glspl{pinn}, and propose \glspl{lpinn} in \cref{sec:pinn} and \cref{sec:lpinn}, respectively.
These architecture are compared with numerical experiments in \cref{sec:experiments}, and the conclusions is followed in
\cref{sec:conclusions}.

\section{Kolmogorov \nwidth\ of the Failure Modes of \glspl{pinn}}
\label{sec:problem}
In this section, we connect the dimensionality of the problem to the training difficulties in \glspl{pinn}.
Specifically, we provide numerical evidence to connect the decay of singular values of the snapshots to the difficulties in the training phase.
This measure corresponds to transport phenomena, convection, traveling waves, and moving fronts.
Although limited attempts were made to quantify connection of dimensionality and slow convergence of some specific classes of \gls{ml} architectures~\cite{E_RMS_2021, Wojtowytsch_IEEEAI_2020}, many of the questions regarding the choice of activation functions, norms, and architecture remains open~\cite{E_RMS_2021}.
Formal investigations of such questions are out of the scope of the present \doc.

In approximation theory, Kolmogorov \nwidth~is a measure of how close
\ndimensional\ subspaces can approximate the solution 
manifold, \manifold~\cite{Quarteroni_2016}.
The following definitions briefly explains this measure~\cite{Pinkus_1985}.

{\textbf{Definition}: 
Let $\manifold$ be a normed linear space and $\reducem{\manifold}$ any 
n-dimensional subspace of $\manifold$. For each $x \in \manifold$, $\delta 
\left(x, \reducem{\manifold}\right)$ shall denote the distance of the 
n-dimensional subspace $\reducem{\manifold}$ from $x$, defined by
\begin{equation}\label{eqn:deltanM}
\delta \left(x; \reducem{\manifold}\right) =  
\inf \left\{ \| x - y \|_{X} : y \in \reducem{\manifold} \right\}.
\end{equation}
If there exist a $y^* \in \reducem{\manifold}$ for which $\delta \left(x, 
\reducem{\manifold}\right) =  \| x - y^* \| $, then $y^*$ is the best 
approximation of $x$ from \reducem{\manifold}. Extending the concept from a 
single element of $x$ to \manifoldS, a given subset of \manifold, the deviation 
of \manifoldS~from \reducem{\manifold}~is defined as
\begin{equation}\label{eqn:deltanMS}
\delta \left(\manifoldS; \reducem{\manifold}\right) =  
\sup_{x \in \manifoldS}
\inf_{y \in \reducem{\manifold}} \| x - y \|,
\end{equation}
representing the worst element of $x \in \manifoldS$ approximated in 
\reducem{\manifold}.
}

{\textbf{Definition}: 
Kolmogorov \nwidth~of \manifold, $d_n\left(\manifold\right)$, is defined as
\begin{equation}\label{eqn:dnM}
d_n\left(\manifoldS; \manifold \right)  :=
 \inf_{\reducem{\manifold}} \delta \left( \manifoldS; 
 \reducem{\manifold}\right),
\end{equation}
where the infimum is taken over all \ndimensional\ subspaces 
(\reducem{\manifold}) of the state space, \manifold.
}

{In the context of Petrov--Galerkin projection schemes, \nwidth\ correlates with the best achievable rate of convergence for a given snapshots~\cite{Melenk_JMAA_2000}. 
The connection between \gls{svd} of the Hankel 
operator and the Kolmogorov \nwidth\ is rigorously 
established~\cite{Djouadi_CDC_2008, Djouadi_PACC_2010}.
It is shown that Kolmogorov \nwidth\  is equivalent to the $(n + 1)^{\textit{st}}$ Hankel singular value~\cite{Unger_ACM_2019}. 
The aforementioned studies strictly connected the singular values of the snapshots to convergence of linear subspaces. 
The extension of such results to non--linear manifolds, such as those discovered through training of \glspl{pinn}, has not been established yet. 
However, we numerically evaluate whether the rate of decay of singular values could be used as a simple \apriori\ guideline to the complexity of the training phase of \glspl{pinn}.
}

\begin{figure}[t!]
	\centering
	\subfloat[$\convspeed=0$]
	{
		\label{fig:a0}
		\includegraphics[scale=\myscalethree]{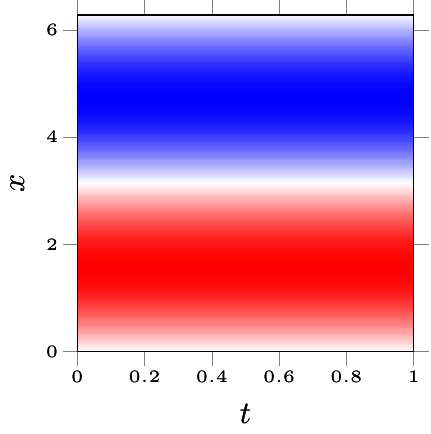}
	}
	\subfloat[$\convspeed=50$]
	{
	\label{fig:a50}
	\includegraphics[scale=\myscalethree]{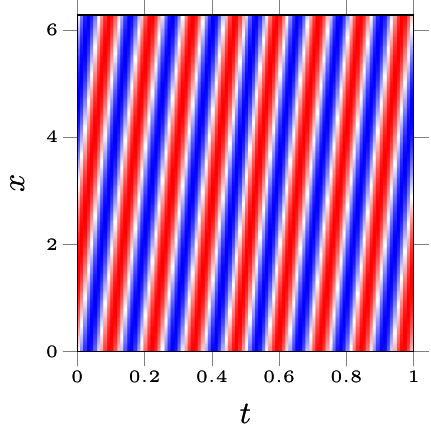}
	}
	\subfloat[Singular value spectra]
	{\label{fig:b1}
		\includegraphics[scale=\myscalethree]{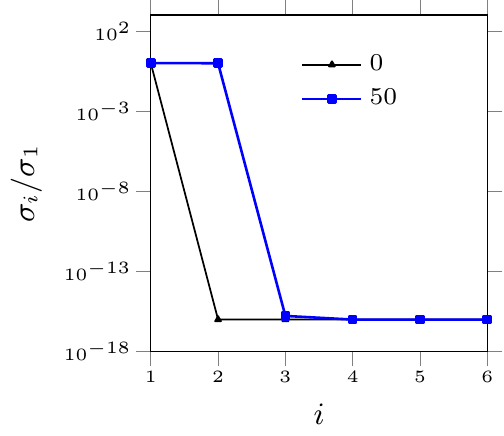}
	}
	\caption{Convection equation. 
		In \labelcref{fig:a0} and \labelcref{fig:a50}, dark red and dark blue represent $+1$ and $-1$, respectively.
		In \labelcref{fig:b1}, the black triangle and blue rectangle markers represent $c=0$ (success mode) and $c=50$ (failure mode), respectively.}
	\label{fig:conv_sin_svd}
\end{figure}

\begin{figure}[t!]
	\centering
	\subfloat[$\rho=1$]
	{\label{fig:reaction0}
		\includegraphics[scale=\myscalethree]{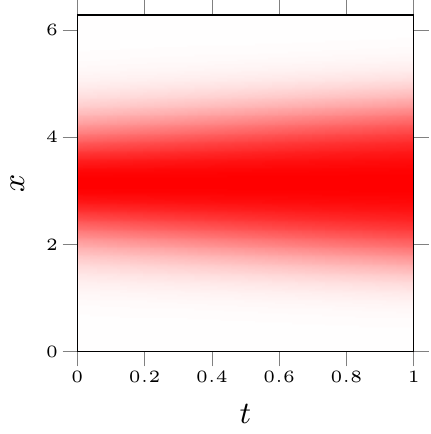}
	}
	\subfloat[$\rho=10$]
	{\label{fig:reaction10}
		\includegraphics[scale=\myscalethree]{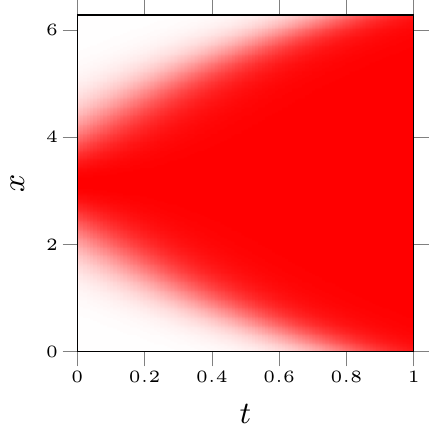}
	}
	\subfloat[Singular value spectra]
	{\label{fig:reactionsvd}
		\includegraphics[scale=\myscalethree]{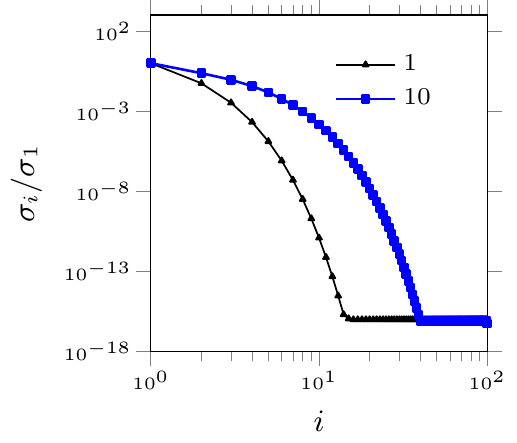}
	}
	\caption{Reaction equation. 
	In \labelcref{fig:reaction0} and \labelcref{fig:reaction10}, dark red and white represent $+1$ and $0$, respectively.
	In \labelcref{fig:reactionsvd}, the black triangle and blue rectangle markers represent $\rho=1$ (success mode) and $\rho=10$ (failure mode), respectively.}
	\label{fig:reaction_svd}
\end{figure}

\begin{figure}[t!]
	\centering
	\subfloat[$\left( \viscosity,\rho \right)=\left(2,5\right)$]
	{\label{fig:reactiondiffusion0}
		\includegraphics[scale=\myscalethree]{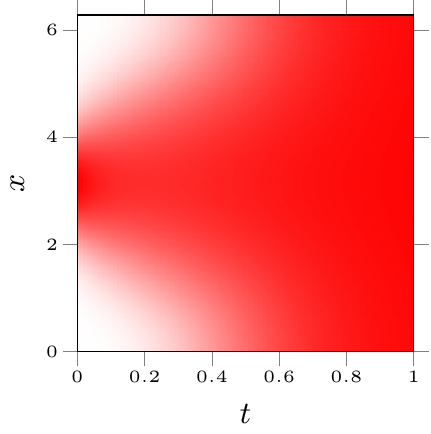}
	}
	\subfloat[$\left( \viscosity,\rho \right)=\left(6,5\right)$]
	{\label{fig:reactiondiffusion10}
		\includegraphics[scale=\myscalethree]{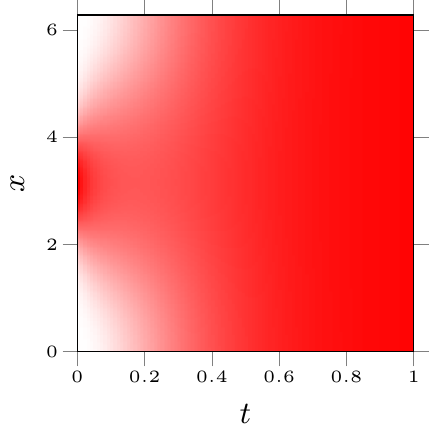}
	}
	\subfloat[Singular value spectra]
	{\label{fig:reactiondiffusionsvd}
		\includegraphics[scale=\myscalethree]{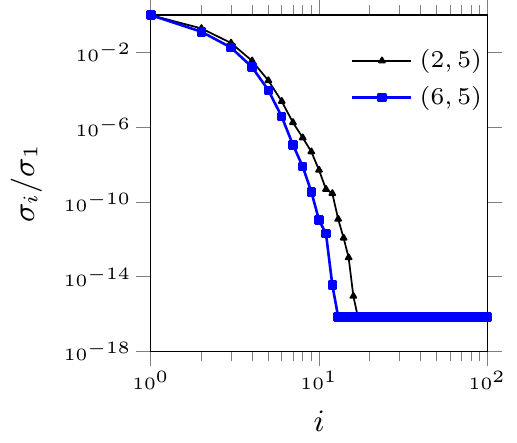}
	}
	\caption{Reaction--diffusion equation. 
	In \labelcref{fig:reactiondiffusion0} and \labelcref{fig:reactiondiffusion10}, dark red and white represent $+1$ and $0$, respectively.
	In \labelcref{fig:reactiondiffusionsvd}, the black triangle and blue rectangle markers represent $\left(\nu,\rho\right)=\left(5,2\right)$ (failure mode) and $\left(\nu,\rho\right)=\left(6,5\right)$ (failure mode), respectively.}
	\label{fig:reactiondiffusion0_svd}
\end{figure}

\begin{figure}[htbp]
	\centering
	\subfloat[$\convspeed=0$]
	{\label{fig:burgers0}
		\includegraphics[scale=\myscalethree]{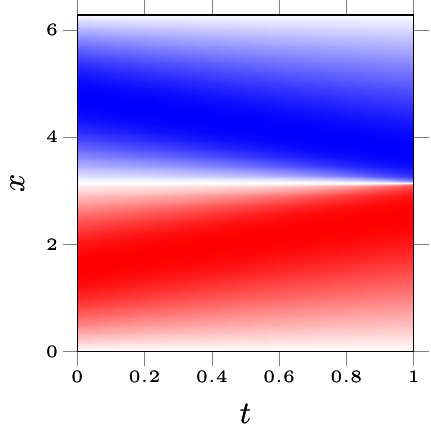}
	}
	\subfloat[$\convspeed=50$]
	{\label{fig:burgers50}
		\includegraphics[scale=\myscalethree]{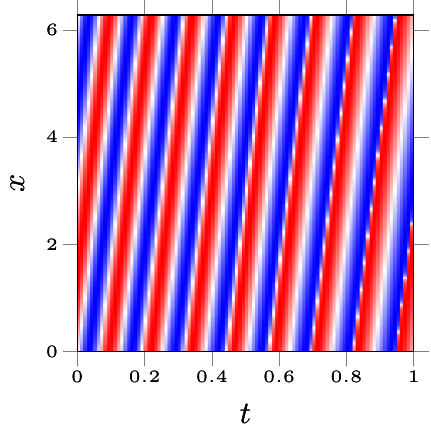}
	}
	\subfloat[Singular value spectra]
	{\label{fig:burgerssvd}
		\includegraphics[scale=\myscalethree]{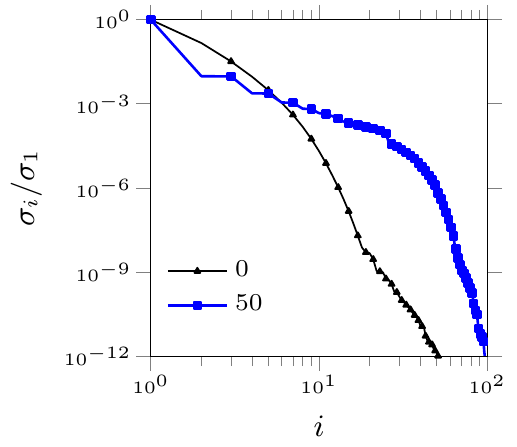}
	}
	\caption{Burgers' equation. 
	In \labelcref{fig:burgers0} and \labelcref{fig:burgers50}, dark red and dark blue represent $\convspeed+1$ and $\convspeed-1$, respectively.
	In \labelcref{fig:b1}, the black triangle and blue rectangle markers represent $\convspeed=0$ (success mode) and $\convspeed=50$ (failure mode), respectively.}
	\label{fig:burgers_svd}
\end{figure}

In \crefrange{fig:conv_sin_svd}{fig:burgers_svd}, the snapshots of the solution and the corresponding singular value decays are plotted for given snapshots of convection equation,
\begin{equation} \label{eqn:conv0}
	\frac{\partial w}{\partial t} 
	- \convspeed \frac{w}{\partial x} = 0,
\end{equation}
reaction equation,
\begin{equation} \label{eqn:reaction}
	\frac{\partial w}{\partial t} -\rho w \left(1-w\right) =0,
\end{equation}
reaction--diffusion equation, 
\begin{equation} \label{eqn:reaction_diffusion}
	\frac{\partial w}{\partial t}-\nu \frac{\partial^{2} w}{\partial x^{2}}-\rho w(1-w) =0,
\end{equation}
and Burgers' equation,
\begin{equation} \label{eqn:burgers0}
	\frac{\partial w}{\partial t} 
	- w \frac{w}{\partial x} = \nu \frac{\partial^2 w}{\partial x^2}.
\end{equation}
where the state variable $w=w(x,t)$ exists in the domain $(x,t) \in [x_a,x_b] \times [0,\tmax]$, and the \glspl{pde} are equipped with initial
conditions $w(x,0) = w_0(x)$, and appropriate boundary conditions at $x_a$, and
$x_b$.
Convection speed, viscosity, and reaction coefficient are denoted by $\convspeed$, $\nu$, and $\rho$, respectively.

The successful and failing regimes of \crefrange{eqn:conv0}{eqn:reaction_diffusion}
were investigated in~\cite{Krishnapriyan_NIPS_2021}. 
We identify Burgers' equation \cref{eqn:burgers0}, in certain regimes, as an additional challenging test case for \glspl{pinn}.
In each of these experiments, the solution of the governing equation is depicted in the space--time domain, and the corresponding singular value spectra are also compared, {\color{\mycolor} where $\sigma_i$ is the $i^{\textit{th}}$ singular value of the snapshots of the solution}.

In \cite{Krishnapriyan_NIPS_2021}, it was shown that \glspl{pinn} for convection become hard to train for $c>10$ (failure modes). Subsequently, the cases with $c=0$ and $c=50$ are compared in \cref{fig:conv_sin_svd}.
It is clear that $\sigma_2/\sigma_1$ increases quickly as $c$ increases.
For the reaction equation, the decay of singular values becomes slower as $\rho$ increases, in a similar trend, the \glspl{pinn} are more difficult to train as $\rho$ increases.
For the cases of reaction--diffusion in \cite{Krishnapriyan_NIPS_2021}, all cases show similar slow rate of decay of singular values, and in a similar trend, training in all the cases encounters difficulties.
Finally, we consider Burgers' equation, a case successfully solved in early studies of \glspl{pinn} \cite{Raissi_JCP_2019}. However, in that case, a viscous shock forms and collapses on its place, without {\it traveling} in the domain.
In this \doc, to impose the shock moving through the domain, the initial condition is offset from zero.
Similar to the convection case, the rate of decay of singular values decreases as the speed of the shock is increased.
Subsequently, similar challenges in training of \glspl{pinn} for Burgers' equation in convection--dominated regimes is observed.
In such cases, the shock travels in the domain before collapsing/diffusing (see \cref{sec:burgers}).

These experiments show that although the \glspl{nn} can express a non--linear manifold, the dimensionality of the solution on the linear optimal subspace of singular vectors can still inform the convergence behavior of \glspl{pinn}, for comparable governing equations.
However, the singular value spectra between two different governing equations do not explain difficulty of the training, e.g., collapsing shock has a slower rate of decay of singular values compared to convection with high speed and yet the training phase is well--behaved. 
Nevertheless, in all cases the network is hard to train at the presence of traveling features, such as shock, fronts, and gradients.

%

\begin{figure}[t!]
	\centering
	\subfloat[]
	{
		\label{fig:synthetic1}
		\includegraphics[]{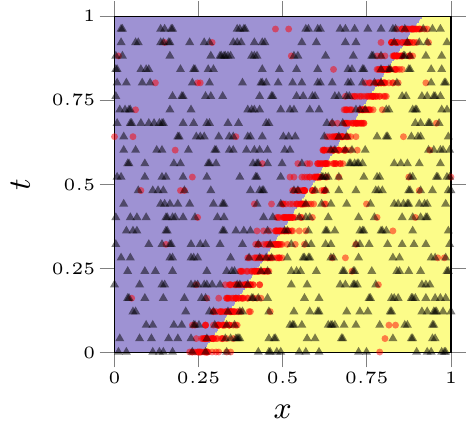}
	}
	\subfloat[]
	{
		\label{fig:synthetic2}
		\includegraphics[trim=0 3 0 0,clip,]{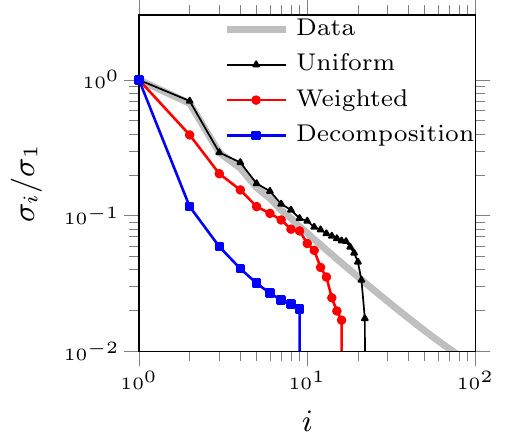}
	}
	\caption{
		a. Synthetic traveling shock and the location of the samples, (i) uniform sampling (black triangles), and (ii) weighted sampling (red circles), 
		b. The corresponding singular value spectra of the
		uniformly sampled data (black triangles), 
		the weighted sampled data (red circles), and
		decomposed data in a limited temporal domain (blue squares). 
	}
	\label{fig:synthetic}
\end{figure}

Some of the successful remedies in \cite{Mao_JCP_2020, Krishnapriyan_NIPS_2021, Bihlo_JCP_2021} can be explained by breaking the Kolmogorov \nwidth, a recognized paradigm in finite element method of solving \glspl{pde} \cite{Evans_CMAME_2009, Mirhosseini_arxiv_2021}, data assimilation \cite{Taddei_SIAM_JSC_2018}, \glspl{nn}--based \glspl{rom} \cite{Mojgani_AAAI_2021, Shady_PoF_2019,Dutta_MCA_2022}, projection-based \glspl{rom} \cite{Mojgani_thesis_2020, Nonino_arxiv_2019, Peherstorfer_SIAM_SC_2020, Rim_arxiv_2020,  Taddei_SIAM_2020,  Shady_Fluids_2020, Mirhosseini_arxiv_2021, Barnett_arxiv_2022, Krah_arxiv_2022,Jie_PRF_2021}, flexDeepONet \cite{Venturi_arxiv_2022},  and projection--based \glspl{rom} on \gls{nn}--based manifolds \cite{Lee_JCP_2020, Kim_JCP_2022}.
To demonstrate the effect of the proposed remedies on the rate of decay of (normalized) singular values, we consider the synthetic data of \cref{fig:synthetic1}, representing a traveling shock.  
The snapshot is constructed on $\numgrid=256$ spatial grid points and $\numtime=500$ time steps.

It is shown that in \glspl{pinn}, accuracy can be increased by over--sampling the traveling shock, in contrast to the uniformly sampled data~\cite{Mao_JCP_2020}.
Here, at each time level $25$ data points are sampled (i) randomly and uniformly, (ii) randomly but weighted where the probability of sampling is proportional to the absolute value of the gradient of the data.
As in \cref{fig:synthetic2}, while sampled data has a lower rank compared to the full data, the first singular values of the data remains unchanged when sampled uniformly (a property used in randomized singular value decomposition). 
However, the rate of decay of the singular values increases by over--sampling the shock.

Parallel--in--time decomposition \cite{Bihlo_JCP_2021} and sequence-to-sequence learning \cite{Krishnapriyan_NIPS_2021} decompose the temporal domain into short time intervals. 
This strategy is similar to \gls{pid} (in linear subspace), applied to \glspl{rom} \cite{Shady_Fluids_2020} and \gls{lstm} networks \cite{Shady_PoF_2019}, and effectively reduces the Kolmogorov \nwidth\ of the data.
Considering the synthetic data, the temporal domain is decomposed to $25$ consecutive time steps, significantly increasing the rate of decay of singular values as in \cref{fig:synthetic2}.
Subsequently, increasing the temporal domain, decreases the rate of decay.
Decomposition of the computational domain, in both space and time, is possible using \glspl{xpinn}, originally developed to tackle the scalibility of \glspl{pinn}~\cite{Jagtap_CCP_2020}.

Given the numerical evidence in this section, one paradigm to tame the training in \glspl{pinn} 
is to reformulate the problem on a manifold such that Kolmogorov \nwidth\ is decreased, or equivalently, the rate of decay of the singular values is increased.
For non--linear convection--diffusion flows, where convection dominates diffusion, such goal is achievable by reformulating the governing equations on the characteristics curves \cite{Mojgani_2017, Lu_JCP_2020}.

\section{PINNs for Non--linear Convection--Diffusion}
\label{sec:convdiff}
Consider the following scalar, one--dimensional convection--diffusion equation
\begin{equation} \label{eqn:E-HFM}
	\residualnull : = \frac{\partial w(x,t)}{\partial t} 
	+ f_1(x,t,w)\frac{\partial w(x,t)}{\partial x}
	- f_2(x,t,w)\frac{\partial^2 w(x,t)}{\partial x^2}
	= 0,
\end{equation}
in the domain $(x,t) \in [x_a,x_b] \times [0,\tmax]$, with initial
conditions $w(x,0) = w_0(x)$, and appropriate boundary conditions at $x_a$ and
$x_b$. 

To address the complexity of the training, the governing
equation \cref{eqn:E-HFM}, is reformulated in the Lagrangian frame of reference
\begin{subequations}\label{eqn:L_HFM}
	\begin{align}
		\residualxnull := \frac{d x}{d t} &-  f_1(x,t,w) = 0,\label{eqn:L_HFM_con_lines}\\
		\residualwnull := \frac{\partial w}{\partial t} &-  f_2(x,t,w)\frac{\partial^2 w}{\partial x^2} = 0 \label{eqn:L_HFM_con_state}.
	\end{align}
\end{subequations}
where $x$ is the characteristic curves and $w$ is the state variable on the  characteristic curves. 
For the sake of simplicity of the notations, we do not differentiate between the state variable on the Eulerian formulation (stationary grid, \cref{eqn:E-HFM}) and the Lagrangian formulation (moving grid, \cref{eqn:L_HFM}). 

\subsection{Traditional PINNs}
\label{sec:pinn}

A \gls{pinn} architecture is composed of a densely connected \gls{ann} that minimizes a loss comprised of the residual equation \cref{eqn:E-HFM} evaluated at the collocation points, data, and the initial and boundary points~\cite{Raissi_JCP_2019}.
The output of the network, is the state parameter, $\bm{\stateparnull}$, 
\begin{equation} \label{eqn:NN}
\bm{\stateparnull} = 
\nntheta{\inputvec} =
				\activationi{\numlayer}{
						\weight{\numlayer}
							\activationi{\numlayer-1}{
								\weight{\numlayer-1}
									\cdots
									\activationi{1}{
									\weight{1} \inputvec
									+\biasi{1}
								} 
								\cdots+\biasi{\numlayer-1}
							} 
						+\biasi{\numlayer}
					} ,
\end{equation}
where the input vector is the concatenation of the spatial and temporal location, i.e., $\inputvec = \transpose{ \left[ \bm{x}_i, t_i \right] } \in \realsetO{\Dim+1}$, where $\Dim$ is the dimension of physical space,  $\activationi{i}{.}$ is the activation function at the $i^{\textit{th}}$-layer,  $\weight{1} \in \realset{\numwidth}{\left(\Dim+1\right)}$, $\weight{i} \in \realset{\numwidth}{\numwidth}, \forall i\in \left\{2,\cdots,\numlayer-1 \right\}$, and $\weight{\numwidth} \in \realset{\numstate}{\numwidth}$ are the weights, and $\biasi{1} \in \realsetO{\numwidth}, \forall i\in \left\{1,\cdots,\numlayer-1 \right\}$, and $\biasi{\numlayer} \in \realsetO{\Dim}$ are biases.
The weights and biases are learned to minimize the so--called physics informed loss, minimizing the residual equation and the appropriate boundary and initial conditions, i.e., 
\begin{equation} \label{eqn:cost}
\loss =  \lossresidual +  \lossbc +  \lossic,
\end{equation}
where
\begin{subequations}\label{eqn:costdefine}
\begin{align}
\lossresidual &= \penaltyr \fracsum{\Nr} \NormvecT{	\residual{\bm{x}_i,t_i}	}, \label{eqn:lossr} \\
\lossbc &= \penaltybc \fracsum{\Nbc} \NormvecT{	\bc \left[\bm{\statepar}\right] \left(\bm{x}_{bc}^i, t_bc^i\right) },
\label{eqn:lossbc} \\
\lossic &= \penaltyic \fracsum{\Nic} \NormvecT{\statepar\left(\bm{x}, 0\right) -
				  \ic \left(\bm{x}_{ic}^i\right)		},
\label{eqn:lossic}
\end{align}
\end{subequations}
and $\scriptstyle \left\{t_{r}^i, \bm{x}_{ic}^i\right\}_{i=1}^{\Nr}$
is the set of temporal and spatial coordinates of the collocation points where the residual is evaluated, and 
$\scriptstyle \left\{\bm{x}_{ic}^i\right\}_{i=1}^{\Nic}$, is the set of coordinates where the initial condition is known ($\ic \left(\bm{x}_{ic}^i\right)$ at $t=0$), and 
$\scriptstyle \left\{t_{bc}^i, \bm{x}_{ic}^i\right\}_{i=1}^{\Nbc}$ is a set of temporal and spatial coordinates of the boundary points ($	\bc \left[\bm{\statepar}\right] \left(\bm{x}_{bc}^i, t_bc^i\right)$). 
The hyperparameters $\penaltyr$, $\penaltybc$, and $\penaltyic$ are scalars tuned to enhance the convergence.
Augmenting the loss using more data points leads to faster convergence, especially in convection--dominated problems~\cite{Abreu_ICCS_2021}. However, we intentionally refrain from using any data points to evaluate the convergence of \glspl{pinn} as a solver (no--data regime). 
To solve convection--diffusion equation, \cref{eqn:E-HFM} is used as the residual term in~\cref{eqn:lossr}.

In this \doc, without loss in generality, we limit the spatial domain to \gls{1d} space. 
The periodic boundary condition is strictly enforced using a custom layer~\cite{Bihlo_JCP_2021}, and therefore $\penaltybc=0$.
The custom layer in $x \in \left[0, 2 \pi \right]$ transforms the domain to a polar coordinate, i.e.,
\begin{equation}\label{eqn:layer0}
\transformlayer{x} = \transpose{
	\left[ \cos\left(x \right),  \sin\left(x \right) \right]
	},
\end{equation}
and out--put of this layer is fed into the traditional \gls{nn} as described in~\cref{eqn:NN} with appropriate adjustment of the dimension of the weight of the first layer, i.e., $\weight{1} \in \realset{\numwidth}{\left(2\Dim+1\right)}$, increasing the number of network variables by only $\numwidth\times\Dim$. 
Further discussion of strictly enforcing the boundary conditions can be found  in~\cite{Dong_JCP_2021}.

\begin{figure}[!t]
	\centering
	\includegraphics[scale=\myscalearch]{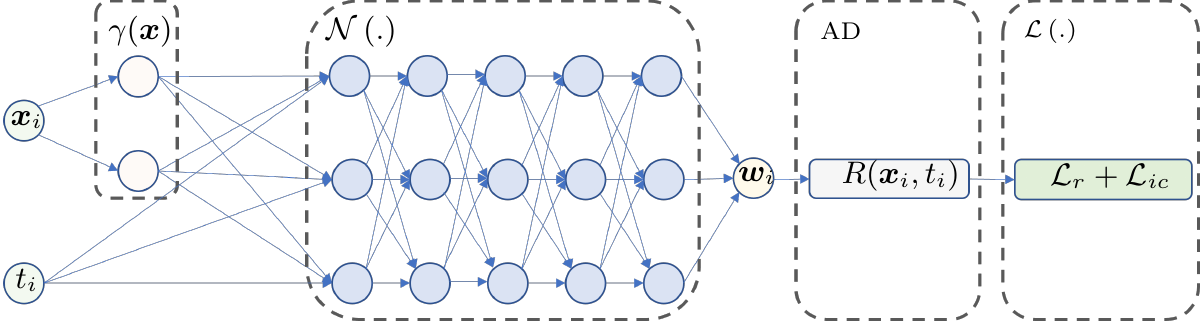}
	\caption{The traditional \glspl{pinn} architecture with periodic boundary condition.}
	\label{fig:archPINN}
\end{figure}

\subsection{Proposed Lagrangian PINNs}
\label{sec:lpinn}

\begin{figure}[!t]
	\centering
	\includegraphics[scale=\myscalearch]{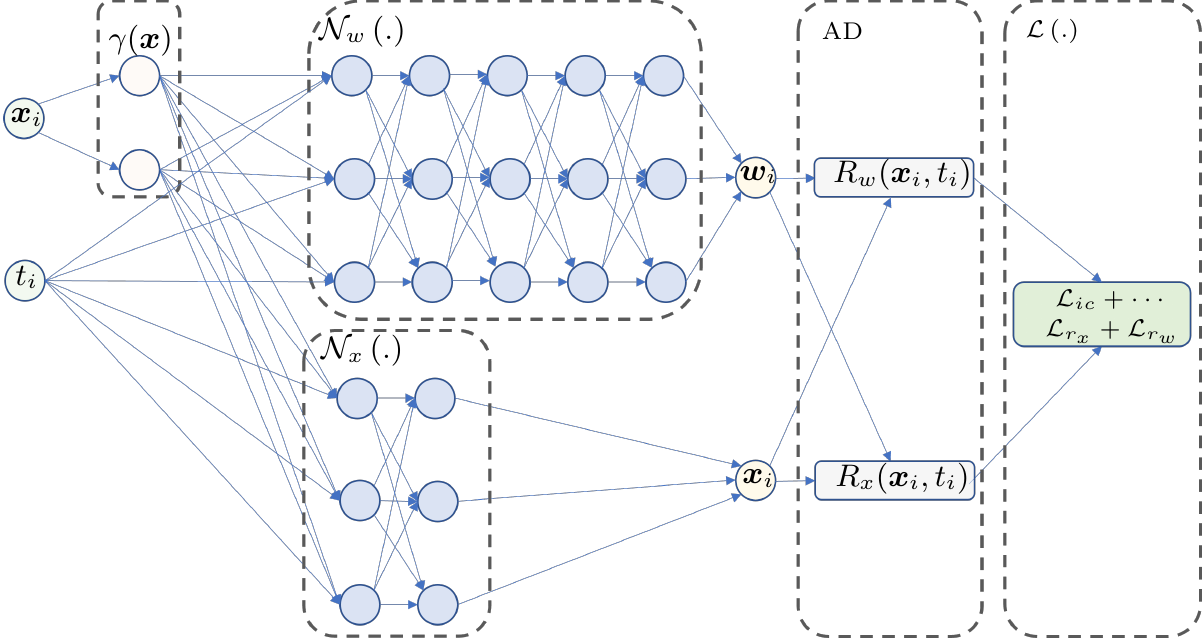}
	\caption{The proposed \glspl{lpinn} architecture with periodic boundary condition.}
	\label{fig:archLPINN}
\end{figure}

In this section, we describe the additional changes to the architecture of the traditional \glspl{pinn} to conform with the Lagrangian formulation to satisfy \cref{eqn:L_HFM}.
We propose a parallel architecture comprised of two branches.
The first branch solves for the characteristics curves and minimizes \cref{eqn:L_HFM_con_lines}, i.e.,
\begin{equation} \label{eqn:LNNx}
\bm{x} = 
\nnthetax{\inputvec} =
\activationi{\numlayer}{
	\weight{\numlayer}
	\activationi{\numlayer-1}{
		\weight{\numlayer-1}
		\cdots
		\activationi{1}{
			\weight{1} \inputvec
			+\biasi{1}
		} 
		\cdots+\biasi{\numlayer-1}
	} 
	+\biasi{\numlayer}
},
\end{equation}
and the second branch solves for the state parameter on the characteristics curves, and minimizes \cref{eqn:L_HFM_con_state}, i.e.,
\begin{equation} \label{eqn:LNNu}
	\bm{\stateparnull} = 
	\nnthetau{\inputvec} =
	\activationi{\numlayer}{
		\weight{\numlayer}
		\activationi{\numlayer-1}{
			\weight{\numlayer-1}
			\cdots
			\activationi{1}{
				\weight{1} \inputvec
				+\biasi{1}
			} 
			\cdots+\biasi{\numlayer-1}
		} 
		+\biasi{\numlayer}
	},
\end{equation}
where all the parameters are defined similar to the network in \cref{sec:pinn}. 
The two branches can be of different width, and depth.
The output of the network is the state parameter on the characteristics curves minimizing the loss, i.e.,
\begin{equation} \label{eqn:cost_Lag}
\loss =  \lossresidualx +  \lossresidualw +  \lossic,
\end{equation}
where $\lossresidualx$ and $\lossresidualx$ are the residual associated with~\cref{eqn:L_HFM}, and $\lossic$ is the loss associated with initial condition of both the state and grid.
Finally, one can interpolate the states from the Lagrangian to the Eulerian frame of reference, $\stateparinterp$, if necessary. 
The proposed architecture is depicted in \cref{fig:archLPINN}.

We recognize the residual equations in \cref{eqn:L_HFM} can also be minimized in an architecture similar to that of the traditional \glspl{pinn}. 
However, the proposed two--branch architecture leverages the inherent low--dimensionality of the characteristics~\cite{Mojgani_2017}, to build a shallow and efficient network to solve \cref{eqn:L_HFM_con_lines}.

\section{Experimental Results}
\label{sec:experiments}
In this section, the traditional \gls{pinn} and the proposed \gls{lpinn} are compared. 
The spatio--temporal grid is of size $\left(\numgrid,\numtime\right)=\left(2^{8},100\right)$ for $(x,t) \in [0,2\pi] \times [0,1]$.
In the convection and convection--diffusion cases, the initial condition is $\stateparinit = \sin \left(2 \pi x\right)$, and the numerical solutions in \cite{Krishnapriyan_NIPS_2021} are considered as the truth.
In the the Burgers' case, the initial condition is 
$\stateparinit = \sin \left(2\pi x\right)+\convspeed$, and solution of a Fourier pseudo--spectral solver is considered as the truth.

In all cases, the \glspl{pinn} have $4$ hidden-layers, and the  \glspl{lpinn} have an additional shallow branch with $2$ hidden-layers. 
All activation functions are $\activation(.)=\tanh(.)$, except those of the last layer, where activation is linear (identity).
The hyper--parameters in \cref{eqn:cost} are $\penaltybc=0$, $\penaltyic=1000$, and $\penaltyr=10$.
Adam optimizer~\cite{Kingma_arxiv_2014} with $10^5$ iterations and learning rate of $0.01$ are used for all the cases.

The error is defined as,
\begin{equation} \label{eqn:error}
\text{Error} = \frac{\NormT{\statepar - \stateparinterp}}{\NormT{\statepar}},
\end{equation}
where $\statepar$ is the truth, and $\stateparinterp$ is the output of the network (interpolated) on the Eulerian grid, both after removing the boundary condition.
A quadratic scheme is used to interpolate the output of \gls{lpinn} to the Eulerian grid.

\subsection{Convection}
\label{sec:conv}

\begin{figure}[t!]
	\centering
	\includegraphics[scale=\myscale]{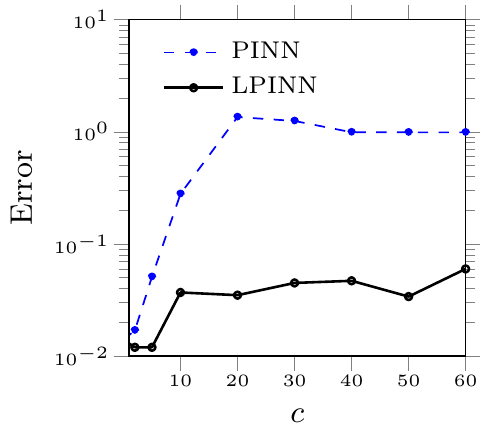}
	\caption{Comparison of the error in \gls{pinn} (dashed blue) vs. the proposed \glspl{lpinn} (solid back) for the convection equation \cref{eqn:conv}.}
	\label{fig:convection}
\end{figure}

Consider the inviscid convection equation,
\begin{equation} \label{eqn:conv}
	\frac{\partial w(x,t)}{\partial t} 
	- \convspeed \frac{w(x,t)}{\partial x} = 0,
\end{equation}
and its reformulation in the Lagrangian frame of reference,
\begin{subequations}\label{eqn:conv_Lag}
	\begin{align}
		\frac{d x}{d t} &= \convspeed, \label{eqn:conv_Lag_lines}\\
		\frac{\partial w}{\partial t} &=   0 \label{eqn:conv_Lag_state}.
	\end{align}
\end{subequations}
The solution to \cref{eqn:conv_Lag} is straightforward.
\Cref{eqn:conv_Lag_state} dictates the grid points to move with the constant convection velocity, $\convspeed$, 
while the state variable remains constant along the moving points, \cref{eqn:conv_Lag_state}.
The accuracy of \gls{pinn} and \gls{lpinn} are compared for different convection velocity in \cref{fig:convection}. 
Similar to \cite{Krishnapriyan_NIPS_2021}, the error of \gls{pinn} increases for larger values of $\convspeed$, such that for $c\ge20$, the \gls{pinn} cannot be trained.
In case of the proposed \gls{lpinn}, where the problem is simply reformulated on the Lagrangian frame of reference, the error for all cases remains below $5\%$.
Note that the reported error is also comprised of the error originating from interpolating the predicted state from the moving grid of the Lagrangian frame to the stationary grid of the Eulerian frame.

\begin{figure}[t!]
	\centering
	\def\fileLoc{data/landscape/convection/dense}
	\def\myscale{0.6}
	\subfloat[\gls{pinn}, $\convspeed = 0$]
	{
		\label{fig:landscape_conv_PINN_0}
		\includegraphics[trim={0 0 0 2cm},clip, scale=\myscale]{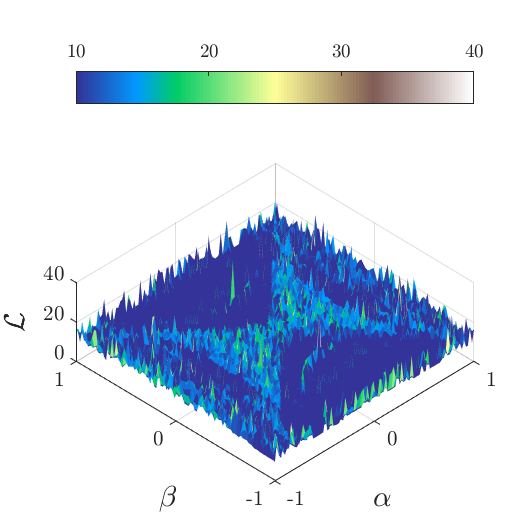}
	}
	\subfloat[\gls{pinn}, $\convspeed = 30$]
	{
		\label{fig:landscape_conv_PINN_30}
		\includegraphics[trim={0 0 0 2cm},clip,scale=\myscale]{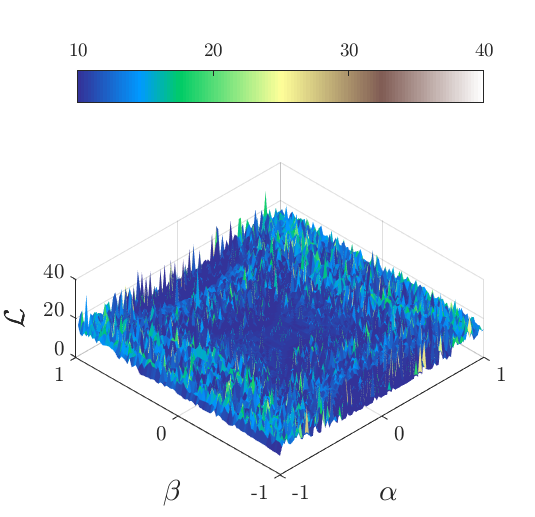}
	}
	\subfloat[\gls{pinn}, $\convspeed = 50$]
	{
	\label{fig:landscape_conv_PINN_50}
	\includegraphics[trim={0 0 0 2cm},clip,scale=\myscale]{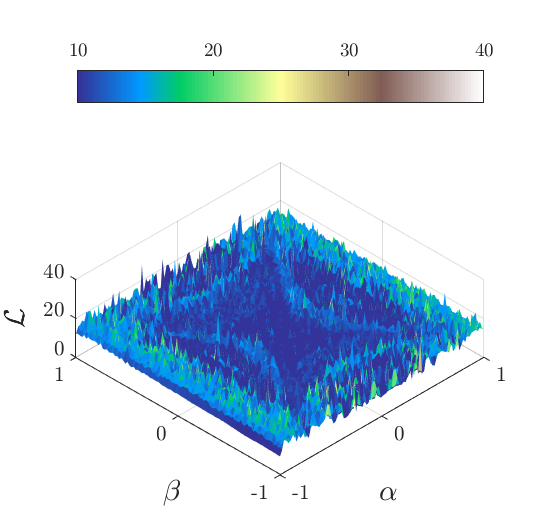}
	}
	\\
	\subfloat[\gls{lpinn}, $\convspeed = 0$]
	{
		\label{fig:landscape_conv_LPINN_0}
		\includegraphics[trim={0 0 0 2cm},clip,scale=\myscale]{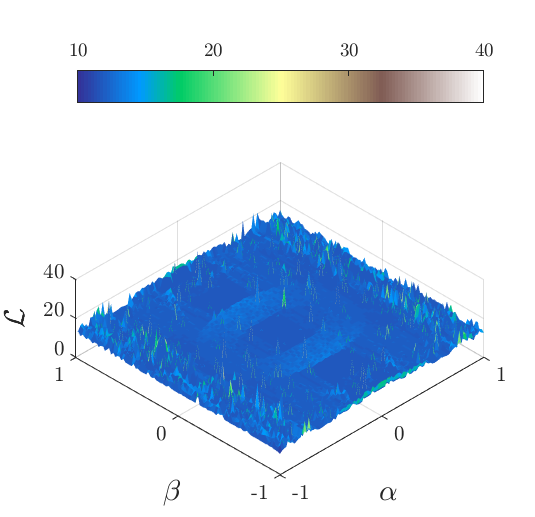}
	}
	\subfloat[\gls{lpinn}, $\convspeed = 30$]
	{
		\label{fig:landscape_conv_LPINN_30}
		\includegraphics[trim={0 0 0 2cm},clip,scale=\myscale]{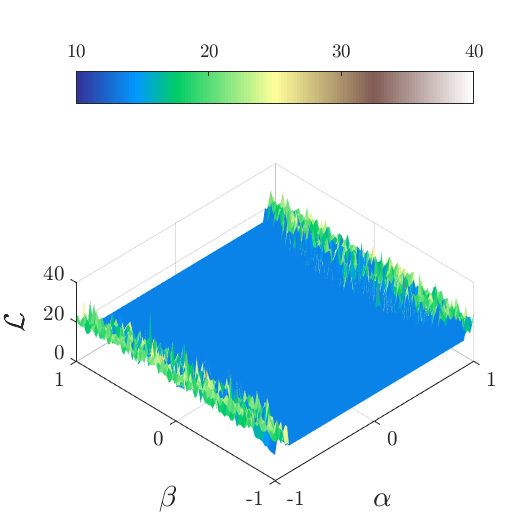}
	}
	\subfloat[\gls{lpinn}, $\convspeed = 50$]
	{
		\label{fig:landscape_conv_LPINN_50}
		\includegraphics[trim={0 0 0 2cm},clip,scale=\myscale]{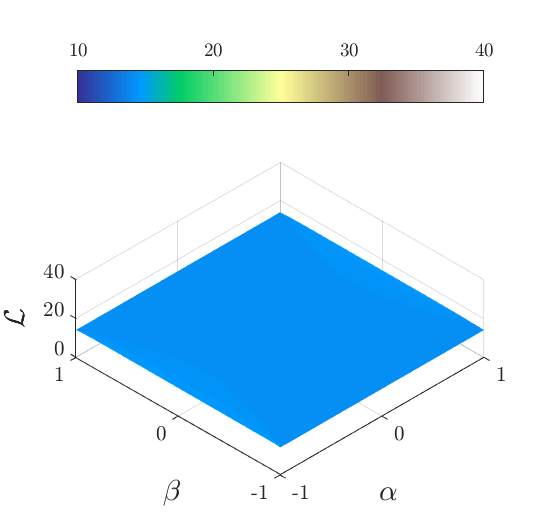}
	}
	\caption{
		The ($\log$ of) loss landscape of convection equation given different convection speeds, $\convspeed \in \left\{0,30,50\right\}$. 
		a-c. \gls{pinn}, 
		d-e. \gls{lpinn}.
	}
	\label{fig:landscape_conv}
\end{figure}

To evaluate the optimality of the trained network, the loss landscape of the network at the end of the training phase is often used as a descriptive measure \cite{Fuks_JMLMC_2020, Krishnapriyan_NIPS_2021, Rohrhofer_arxiv_2022, Basir_SCITECH_2022}.
To compute the loss landscape, the two dominant eigenvectors of the Hessian of the loss with respect to the trainable parameters of the networks, $\bm{\delta}$ and $\bm{\eta}$, are computed using the code provided in \cite{Yao_IEEE_2020}.
Subsequently, the network is perturbed along the eigenvectors and its loss, $\loss'$, is evaluated, i.e., 
\begin{equation} \label{eqn:loss_landscape}
\loss'\left( \alpha , \beta \right) = \loss \left(\NNparam + \alpha \bm{\delta} + \beta \bm{\eta} \right),
\end{equation}
where $\left(\alpha, \beta \right) \in \left[-\alpha_0,\alpha_0\right] \times \left[-\beta_0,\beta_0\right]$.
Finally, $\log{\left( \loss'\left( \alpha , \beta \right) \right)}$ is visualized in \cref{fig:landscape_conv}, for $\convspeed = \left\{0,30,50\right\}$ and for both \gls{pinn} and \gls{lpinn} architectures.
In \cref{fig:landscape_conv_PINN_0}, we recover the saddle shape of the loss landscape for small convection speed as reported for \gls{pinn} in \cite{Krishnapriyan_NIPS_2021}.
Similarly, by increasing $c$, the landscape becomes less smooth (sharper, or more rugged), implying the trained network is not at a minimizer (\crefrange{fig:landscape_conv_PINN_30}{fig:landscape_conv_PINN_50}). 
In the case of \gls{lpinn} (\crefrange{fig:landscape_conv_LPINN_0}{fig:landscape_conv_LPINN_50}), the loss landscapes are significantly smoother compared to their \gls{pinn} counterparts (\crefrange{fig:landscape_conv_PINN_0}{fig:landscape_conv_PINN_50}).
Moreover, the landscape is smooth (flat), even at high $c$, increasing the confidence that the obtained minimizer is a global one.

\subsection{Convection-diffusion}
\label{sec:conv-diff}
\newcommand{\fone}{\ensuremath{ \convspeed }}
\newcommand{\ftwo}{\ensuremath{ \nu }}

\begin{figure}[tb!]
	\centering
	\subfloat[$\nu=1.0$]
	{
	\centering
	\label{fig:condiff_nu=1d00}
	\includegraphics[scale=\myscale]{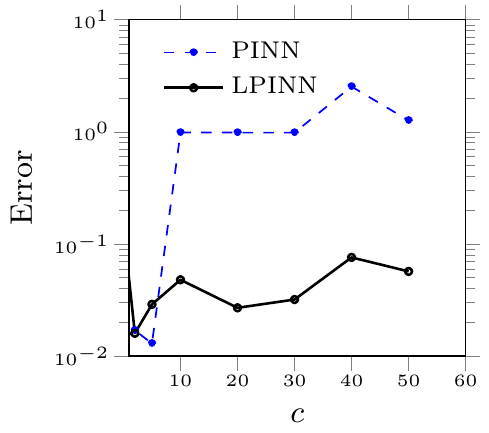}
	}
	\subfloat[$\nu=0.1$]
	{
	\centering
	\label{fig:condiff_nu=0d10}
	\includegraphics[scale=\myscale]{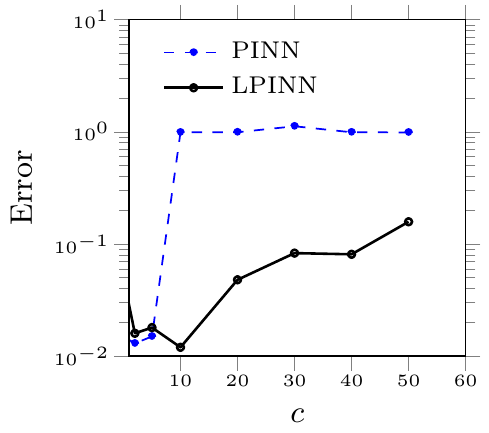}
	}
	\subfloat[$\nu=0.01$]
	{
	\centering
	\label{fig:condiff_nu=0d01}
	\includegraphics[scale=\myscale]{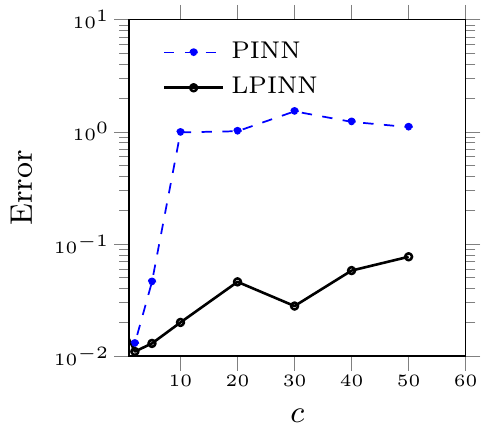}
	}
	\caption{Comparison of the error in \gls{pinn} (dashed blue) vs. the proposed \glspl{lpinn} (solid back) for the convection--diffusion equation \cref{eqn:convdiff} with different $\nu$.}
	\label{fig:condiff}
\end{figure}

Consider the viscous convection--diffusion equation,
\begin{equation} \label{eqn:convdiff}
	\frac{\partial w(x,t)}{\partial t} 
	- \fone \frac{w(x,t)}{\partial x} = \ftwo \frac{\partial^2 w(x,t)}{\partial x^2}  ,
\end{equation}
and its reformulation in the Lagrangian frame of reference,
\begin{subequations}\label{convdiff_Lag}
	\begin{align}
		\frac{d x}{d t} &=  \fone,\label{eqn:convdiff_Lag_lines}\\
		\frac{\partial w}{\partial t} &=   \ftwo \frac{\partial^2 w}{\partial x^2}\label{eqn:convdiff_Lag_state}.
	\end{align}
\end{subequations}
\Cref{fig:condiff} compares the accuracy of \gls{pinn} and the proposed \glspl{lpinn}.
Similar to the inviscid case discussed in \cref{sec:conv}, the error in \glspl{pinn} increases by increasing  $\convspeed$ and they fail to train after a critical $\convspeed$.
Similarly, the \glspl{lpinn} increases by increasing $\convspeed$, however, the error remains around $10 \%$, even in the most challenging cases.

\subsection{Burgers' equation}
\label{sec:burgers}

Consider the viscous Burgers' equation, 
\begin{equation} \label{eqn:burgers}
	\frac{\partial w(x,t)}{\partial t} 
	- w(x,t) \frac{w(x,t)}{\partial x} = \nu \frac{\partial^2 w(x,t)}{\partial x^2}  ,
\end{equation}
and its representation on the Lagrangian frame,
\begin{subequations}\label{burgers_Lag}
	\begin{align}
		\frac{d x}{d t} &=  w(x,t),\label{eqn:burgers_Lag_lines}\\
		\frac{\partial w}{\partial t} &=   \ftwo \frac{\partial^2 w}{\partial x^2}\label{eqn:burgers_Lag_state}.
	\end{align}
\end{subequations}

\begin{figure}[tb!]
	\centering
	\subfloat[$\nu=1.0$]
	{
		\centering
		\label{fig:burgers_nu=1d00}
		\includegraphics[scale=\myscale]{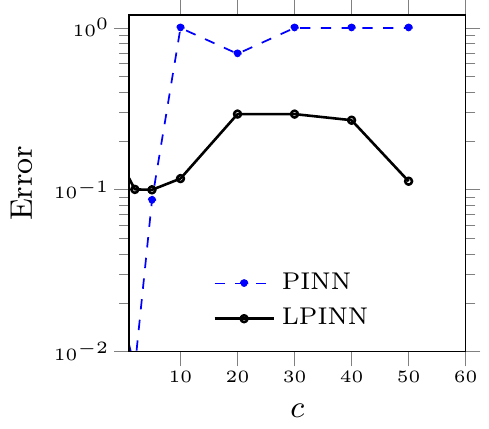}
	}
	\subfloat[$\nu=0.1$]
	{
		\centering
		\label{fig:burgers_nu=0d10}
		\includegraphics[scale=\myscale]{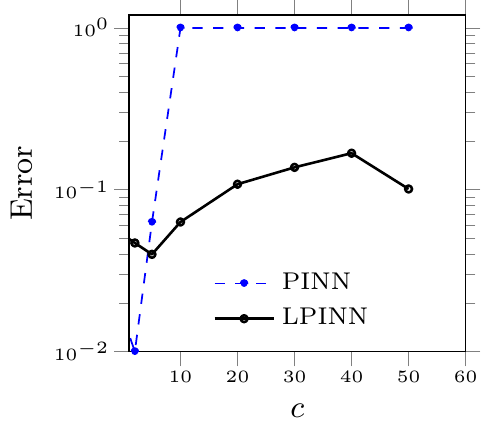}
	}
	\subfloat[$\nu=0.01$]
	{
		\centering
		\label{fig:burgers_nu=0d01}
		\includegraphics[scale=\myscale]{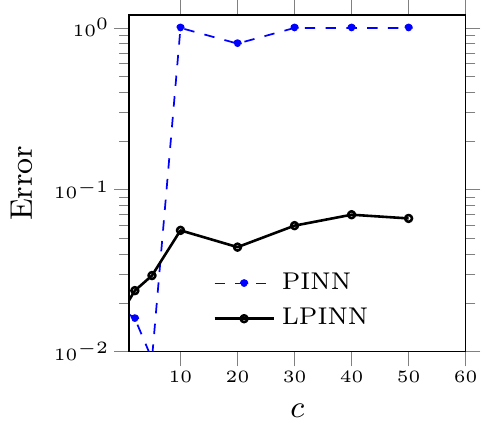}
	}
	\caption{
		Comparison of the error in \gls{pinn} (dashed blue) vs. the proposed \glspl{lpinn} (solid back) for the viscous Burgers' equation \cref{eqn:burgers} with different $\nu$.
	}
	\label{fig:burgers}
\end{figure}

\begin{figure}[tb!]
	\centering
	\subfloat[$\convspeed=30$]
	{
		\centering
		\label{fig:burgers_nu=0d01_C30}
		\includegraphics[scale=\myscale]{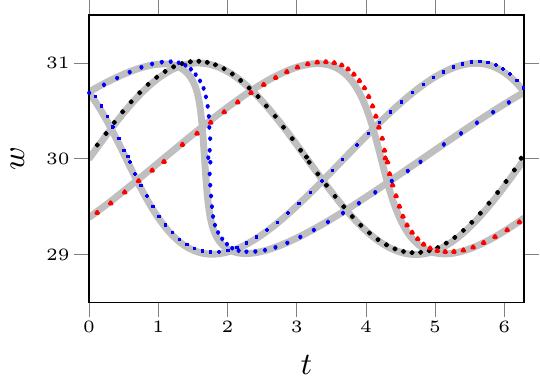}
	}
	\subfloat[$\convspeed=50$]
	{
		\centering
		\label{fig:burgers_nu=0d01_C50}
		\includegraphics[scale=\myscale]{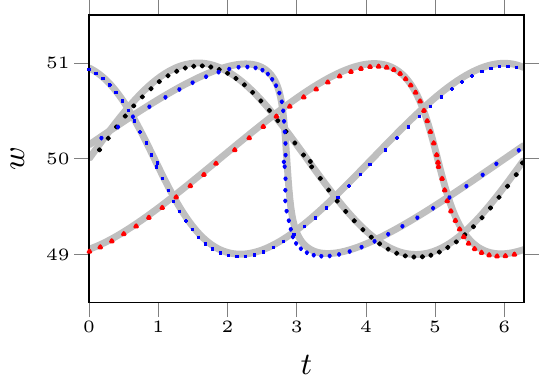}
	}
	\caption{
		Comparison of the proposed \gls{lpinn} with pseudo--spectral solver for the viscous Burgers' equation \cref{eqn:burgers} ($\nu=0.01$) at $t\in\left\{0,\tmax/3, 2\tmax/3, \tmax\right\}$ (black circle, blue square, red triangle, blue diamond) for $\convspeed=\left\{30, 50\right\}$. \Gls{pinn} cannot be trained in both regimes.
	}
	\label{fig:burgers_space_time}
\end{figure}

While the traditional formulation of \gls{pinn} is successfully demonstrated  for Burgers' equation~\cite{Raissi_JCP_2019}, the examined problem lack the main property of challenging cases for training, i.e., the large Kolmogorov \nwidth\ associated with the travel of the shock as in \cref{fig:burgers_svd}.
In \cref{fig:burgers}, and in a similar to the trend in convection--diffusion equation, the \gls{pinn} fails to train for $\convspeed\ge10$, while the \gls{lpinn} is trained for all cases.
The higher error in this case compared to the convection--diffusion equation is due to the higher interpolation error close to the shock.
Note that in these cases the viscosity, $\nu$, is small enough to form the high gradient shock and is large enough to avoid the intersecting characteristics. \cref{fig:burgers_space_time} shows the accuracy of the proposed \gls{lpinn} compared to the numerical solver at different simulation time steps.

\section{Conclusions}
\label{sec:conclusions}

The contributions of this \doc is threefold.

\begin{enumerate}
	\item We described the challenge of training of the traditional architecture of \glspl{pinn} through the lens of approximation theory using Kolmogorov \nwidth, and the rate of decay of singular values of the solution. 
	This realization explains many of the successful remedies in training of \glspl{pinn}.
	More importantly, it can help with identifying unknown challenging problems, and opens of a wide variety of possibility to address them.
	
	\item We identified Burgers' equation in the presence of traveling shocks as another challenging case. 
	The complexity of the training is explained based on our discussion of the irreducibly of the solution on a linear space.
	
	\item We propose \gls{lpinn} for \gls{1d} linear and non--linear convection--diffusion  equations.
	The reformulation of the equations on the characteristics automatically conforms to the direction of travel of the information in the domain, and satisfies the expected causality. 
	More importantly, the solution on the manifold is of lower dimension, i.e., low Kolmogorov \nwidth, and is less sensitive to the system parameters.
	Using the inherent low--dimensionality of the characteristics \cite{Mojgani_2017}, only a shallow branch is added to the traditional \gls{pinn} to minimize the composite loss comprised of residual equations of characteristics, and the state variable on the characteristics. 
	{\color{\mycolor}
	While it is suggested that the condition number of the loss is a probable source of complexity in training of \glspl{pinn} \cite{Krishnapriyan_NIPS_2021}, the proposed \gls{lpinn} architecture is robust with respect to the condition number and show convergence regardless.}
\end{enumerate}

Future work should investigate the problem with shocks where the characteristics intersect.  In such cases, a vanishing viscosity approach or removing the triple point value using Rankine--Hugoniot condition seem to be straightforward solutions~\cite{Leveque_numerical_1992}. 
Similar strategies are even applied to \glspl{pinn} in presence ~\cite{Fraces_arxiv_2021, Abreu_ICCS_2021}.
An extension of the proposed \gls{lpinn} to higher spatial dimensions is possible using Radon transform~\cite{Rim_SIAM_JSC_2018}. 
In cases where characteristics curves are not real, e.g., wave equation, a manifold can be identified by registration--based or feature tracking approaches, e.g.,~\cite{Taddei_SIAM_2020, Taddei_ESIAM_2021, Mojgani_AAAI_2021, Sarna_CMAME_2021, Mirhosseini_arxiv_2021}, and to replace the mass--spring--damper grid models in \gls{ale} formulation of \glspl{pinn}~\cite{Wessels_2020_CMAME}.
Specifically, an optimal and low--rank manifold can be constructed offline by identifying an optimally morphing grid \cite{Mojgani_AAAI_2021}.
Moreover, \gls{lpinn} architecture provides an opportunity for one--shot discovery of an optimal manifold.

%




\section*{Acknowledgements}
This work was supported by an award from the ONR Young Investigator Program (N00014-20-1-2722), a grant from the NSF CSSI program (OAC-2005123). 
Computational resources were provided by NSF XSEDE (allocation ATM170020) and NCAR's CISL (allocation URIC0004).
Our codes and data are available at~\mydepository.

\bibliographystyle{elsarticle-num-names}
\bibliography{library}{}







\end{document}